\documentclass[letterpaper, 10 pt, conference]{ieeeconf}  %

\IEEEoverridecommandlockouts                              %

\usepackage{graphicx}
\usepackage{subcaption}
\usepackage[export]{adjustbox}

\usepackage{amssymb}
\usepackage{amsmath}
\usepackage{commath}
\usepackage{mathtools}

\usepackage[pdfborder={0 0 0}]{hyperref}
\usepackage{import}
\usepackage{paralist}
\usepackage{todonotes}
\usepackage{booktabs} %

\usepackage{siunitx}
\usepackage{microtype}

\title{\LARGE \bf
Physics-Informed Learning for the Friction Modeling \\ of High-Ratio Harmonic Drives}

\author{Ines Sorrentino$^{1,2}$, Giulio Romualdi$^{1}$, Fabio Bergonti$^{1}$, \\ Giuseppe L'Erario$^{1}$, Silvio Traversaro$^{1}$, Daniele Pucci$^{1,2}$%
\thanks{*The paper was supported by the Italian National Institute for Insurance against Accidents at Work (INAIL) ergoCub Project}%
\thanks{${}^{1}$ Artificial and Mechanical Intelligence, Italian Institute of Technology, Genoa, Italy, {\tt\small (e-mail: name.surname@iit.it)}}
\thanks{${}^{2}$  School of Computer Science, University of Manchester, Manchester, UK}%
}

\begin{document}

\maketitle
\thispagestyle{empty}
\pagestyle{empty}

\begin{abstract}
This paper presents a scalable method for friction identification in robots equipped with electric motors and high-ratio harmonic drives, utilizing Physics-Informed Neural Networks (PINN). This approach eliminates the need for dedicated setups and joint torque sensors by leveraging the robot’s intrinsic model and state data. We present a comprehensive pipeline that includes data acquisition, preprocessing, ground truth generation, and model identification. The effectiveness of the PINN-based friction identification is validated through extensive testing on two different joints of the humanoid robot ergoCub, comparing its performance against traditional static friction models like the Coulomb-viscous and Stribeck-Coulomb-viscous models. Integrating the identified PINN-based friction models into a two-layer torque control architecture enhances real-time friction compensation. The results demonstrate significant improvements in control performance and reductions in energy losses, highlighting the scalability and robustness of the proposed method, also for application across a large number of joints as in the case of humanoid robots.

\end{abstract}

\section{Introduction}
\label{sec:introdcution}

Executing tasks requiring complex dynamic motions, such as running, jumping, or navigating uneven terrain, demands careful consideration of the dynamic properties of the low-level joint actuation. Specifically, implementing torque-controlled dynamic motion in bipedal robots using electric actuators is particularly complex, especially in the absence of joint torque sensors. Achieving precise torque control necessitates developing and implementing control algorithms that incorporate accurate dynamic modeling, as well as the identification of motor characteristics and harmonic drive properties. Many humanoid robots, such as ergoCub, iCub3, PAL Robotics' REEM-C, and RoK-3, utilize a combination of motors and high-reduction gears, such as Harmonic Drives, to generate the substantial torques needed for their joints given the robot's weight~\cite{dafarra2024icub3,han2022slope,ossadnik2018adaptive,han2023implementing}. However, high-reduction gears pose significant challenges, including high friction and low backdrivability. This high friction can introduce considerable delays between torque input and output, complicating torque control, especially when rapid responses to dynamically changing torque are required. Furthermore, high friction leads to energy losses, increased wear and tear, and reduced efficiency, ultimately degrading the precision and responsiveness of joint movements. Addressing these challenges requires sophisticated friction models and compensation strategies~\cite{marques2016survey}.

Friction identification can be performed using model-based or learning-based approaches. Model-based approaches, such as Coulomb, viscous, LuGre, Dahl, Leuven, and generalized Maxwell-slip models, are typically categorized into static and dynamic friction models~\cite{awrejcewicz2005analysis}. Static models have limitations in capturing the full spectrum of frictional phenomena and often yield less accurate results than dynamic models~\cite{iurian2005identification}. Moreover, many existing static friction models struggle with discontinuity when transitioning from the pre-sliding to the sliding regime~\cite{olejnik2013application, jin2019joint}.
\looseness=-1
\begin{figure}[t]
\centering
\begin{subfigure}{.49\columnwidth}
  \includegraphics[width=\textwidth]{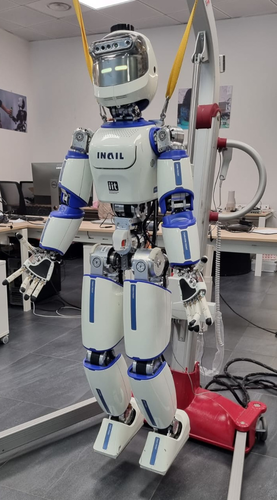}
  \label{fig:ergocub_front}
\end{subfigure}%
\hspace{\fill} %
\begin{subfigure}{.49\columnwidth}
  \includegraphics[width=\textwidth]{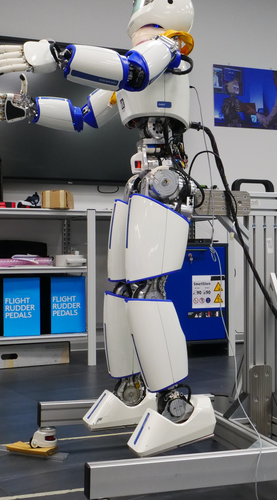}
  \label{fig:ergocub_lat}
\end{subfigure}
\vspace{-20pt}
\caption{The ergoCub humanoid robot.}
\label{fig:ergocub}
\vspace{-20pt}
\end{figure}
The Coulomb-viscous friction model is sufficient for depicting the frictional behavior within multi-body dynamic systems for many applications. For example, the Coulomb-viscous model can estimate and compensate for torque losses during the landing impact of a biped robot~\cite{nagamatsu2017distributed}. The model can be improved by integrating the Stribeck effect, achieving more accurate friction compensation~\cite{jin2019joint}. Nevertheless, traditional static and dynamic friction models typically rely on simplified assumptions and linear approximations, failing to capture the complex system dynamics. In the case of high-ratio harmonic drives, friction exhibits highly non-linear behavior and complex dependencies on various states and conditions, such as the variation of temperature or the presence of backlash, especially during direction changes or low-speed movements~\cite{wolf2018extending}.

\looseness=-1

Models based on Neural Networks (NN) have advantages in modeling complex systems by learning directly from the data. Research on using NN to develop friction models for robot joints is limited, especially for humanoid robots equipped with electric motors and high-ratio harmonic drives. One approach uses a back-propagation NN, optimized by a genetic algorithm, to model static friction in a robot joint~\cite{tu2019modeling}. A second solution models rolling friction using Long Short-Term Memory (LSTM)~\cite{wang2023improved}. However, these approaches require large datasets to ensure diverse and representative training data. Moreover, LSTMs are more computationally intensive than simple feedforward NNs, making it harder to scale for multiple humanoid robot joints while maintaining real-time performance and a low computational load. To overcome these limitations a study identifies the planar frictional torque acting at the contact surface of a double torsion pendulum by adopting a Physics Informed Neural Network (PINN)~\cite{olejnik2023friction}. PINNs overcome problems of small datasets by informatively exploiting a priori knowledge of the non-linear system for constructing robust feedforward NNs. In the physics-informed modeling framework, training of the NN is facilitated by directly embedding the underlying governing physical laws as regularization terms in the cost function of the learning problem.\looseness=-1

Friction identification procedures often require specialized setups, as shown in~\cite{wang2023improved}, where data were collected from an emulated industrial machine axis. However, such setups are not always available, making this technique impractical in many scenarios. An alternative is to collect data directly from the robot, as demonstrated in~\cite{tu2019modeling}, which uses motor angular velocities and joint load torques as inputs for the NN. This method, however, fails when the joint load torque is unknown. When measuring the ground truth is not feasible, friction torque can be computed through dynamic modeling, as in~\cite{olejnik2023friction}. Yet, this approach, tested on a double torsion pendulum, is unsuitable for our problem for two reasons. First, our system incorporates harmonic drives, while the system in~\cite{olejnik2023friction} does not. Second, their method requires two NNs to predict friction torque and angular rotation, which adds computational complexity, making it impractical for humanoid robots with many degrees of freedom.
\looseness=-1

This paper contributes to developing a robust and scalable friction identification method based on Physics-Informed Neural Networks (PINN). The PINN model inputs the history of the position error between joint and motor positions, and joint velocities. This method is specifically tailored for robots with electric motors and high-ratio harmonic drives. Our approach does not require dedicated setups for data acquisition. The pipeline is versatile and applicable to all joints of a humanoid robot, making it neither joint nor robot-dependent. The chosen identification model (PINN) is computationally efficient, enabling scalability to all joints of a humanoid robot while maintaining real-time performance.\looseness=-1

Additionally, we test the PINN model's effectiveness by comparing it with classical static friction models, specifically Coulomb-viscous and Stribeck-Coulomb-viscous. The identification and testing processes are conducted on the humanoid robot ergoCub. The data acquisition procedure involves the accurate selection of exciting trajectories and thorough data processing. We incorporate the estimated friction models into a real-time friction compensation system integrated with a joint torque control architecture~\cite{sorrentino2024ukf}. We demonstrate that the PINN model reduces energy losses to achieve the same high-level tracking performance as the other models.\looseness=-1

The paper is organized as follows. Sec.~\ref{sec:background} introduces the notation and recalls some concepts of floating-base systems, harmonic drive model, an overview of static friction models, and the Physics Informed Neural Network model. Sec.~\ref{sec:methods} details the friction models identification procedure. Sec.~\ref{sec:results} presents the validation results on two diffent joints of the humanoid robot ergoCub. Sec.~\ref{sec:conclusions} concludes the paper.\looseness=-1

\section{Background}
\label{sec:background}

\subsection{Notation}
\begin{itemize}
\item $I_{n}$ and $0_{m \times n}$ denote the $n \times n$ identity matrix and the $m \times n$ zero matrix respectively.
\item $\mathcal{I}$ denotes the inertial frame.
\item $\prescript{\mathcal{I}}{}{p}_\mathcal{B}$ is a vector connecting the origin of frame $\mathcal{I}$ and the origin of frame $\mathcal{B}$ expressed in frame $\mathcal{I}$.
\item Given $\prescript{\mathcal{I}}{}{p}_\mathcal{B}$ and $\prescript{\mathcal{B}}{}{p}_\mathcal{C}$,  $\prescript{\mathcal{I}}{}{p}_\mathcal{C} = \prescript{\mathcal{I}}{}{R}_\mathcal{B} \prescript{\mathcal{B}}{}{p}_\mathcal{C} + \prescript{\mathcal{I}}{}{p}_\mathcal{B}$.
\item $i_m \in \mathbb{R}^n$ is the vector of motor currents with $n$ the number of motors. 
\item $\scalebox{0.8}{$\prescript{}{}{\mathrm{f}}_\mathcal{B}^ \top = \begin{bmatrix} {{f}}_\mathcal{B} ^ \top & {\mu}_\mathcal{B}^ \top \end{bmatrix}$}$ is the wrench acting on a point of a rigid body expressed in the frame $\mathcal{B}$.
\item Given $\prescript{}{}{\mathrm{f}}_\mathcal{B}$ and $\prescript{}{}{\mathrm{f}}_\mathcal{C}$, $\prescript{}{}{\mathrm{f}}_\mathcal{C} = \prescript{\mathcal{C}}{}{X}_\mathcal{B} \prescript{}{}{\mathrm{f}}_\mathcal{B}$, where $\prescript{\mathcal{C}}{}{X}_\mathcal{B} \in \mathbb{R}^{6 \times 6}$ is the adjoint matrix defined as 
\scalebox{0.8}{$\prescript{\mathcal{C}}{}{X}_\mathcal{B} = \small{\begin{bmatrix}
\prescript{\mathcal{C}}{}{R}_\mathcal{B} & \prescript{\mathcal{C}}{}{p}_\mathcal{B}^{\wedge} \prescript{\mathcal{C}}{}{R}_\mathcal{B} \\
0_{3 \times 3} & \prescript{\mathcal{C}}{}{R}_\mathcal{B}
\end{bmatrix}}$}.
\end{itemize}
\looseness=-1

\subsection{Humanoid Robot Model}
A humanoid robot is a floating base multibody system composed of $n+1$ links connected by $n$ joints with one degree of freedom each. The joint positions $s$ and the homogeneous transformation from the inertial frame to the robot base frame $\mathcal{B}$ define the robot configuration. The configuration is identified by the triplet $q = (\prescript{\mathcal{I}}{}{p}_\mathcal{B}, \prescript{\mathcal{I}}{}{R}_\mathcal{B}, s) \in  \mathbb{R}^3 \times SO(3) \times \mathbb{R}^n$. The velocity of the system is characterized by the set $\nu = (\prescript{\mathcal{I}}{}{\dot{p}}_\mathcal{B}, \prescript{\mathcal{I}}{}{\omega}_\mathcal{B}, \dot{s})$ where $\prescript{\mathcal{I}}{}{\dot{p}}_\mathcal{B}$ is the linear velocity of the base frame expressed into the inertial frame, $\prescript{\mathcal{I}}{}{\omega}_\mathcal{B}$ is the angular velocity of the base frame expressed into the inertial frame, and $\dot{s}$ is the time derivative of the joint positions.
\looseness=-1

The equation of motion of a multibody system is described by applying the \textit{Euler-Poincar\'e Formalism}
\begin{equation}
M(q) \dot \nu + h(q, \nu) = B \tau + \sum_{k} {J_k(q)^\top f_{ext,k}} \; ,
\label{eq:robotdynamics}
\end{equation}
where \mbox{$M \in \mathbb{R}^{(6+n) \times (6+n)}$} is the mass matrix, \mbox{$h(q, \nu) \in \mathbb{R}^{6+n}$} accounts for the Coriolis, centrifugal and gravitational effects, $\tau \in \mathbb{R}^{n}$ are the joint torques, $B \in \mathbb{R}^{(6+n) \times n}$ is the selection matrix, $f_{ext,k} \in \mathbb{R}^6$ is the $k$-th contact wrench expressed in the contact frame, $J_k(q)$ is the Jacobian associated to the $k$-th contact wrench~\cite{featherstone2014rigid}.
Humanoid robots actuated by brushless direct current (DC) motors are equipped with high-ratio gearboxes to generate the substantial torques required to move the robot joints, given the robot weight. The dynamics of a DC motor with a harmonic drive is described by the equation $k_{t} i_m = J_m \ddot{\theta} + \frac{1}{r} \tau_F + \frac{1}{r} \tau$
where $k_t$ is the torque constant, $i_m$ is the applied motor current, $J_m \ddot{\theta}$ is the angular acceleration of the motor shaft, $\tau_F$ is the friction torque, $\tau$ is the torque applied to the load, and $r$ represents the reduction ratio~\cite{taghirad1996experimental}. In scenarios with high friction, high reduction ratios, low motor inertia, or low angular acceleration, the frictional effects dominate the system dynamics over the inertial effects of the motor, and $J_m \ddot{\theta} \ll \frac{1}{r} \tau_F$. The joint torque can be simplified in
\begin{equation}
\label{eq:HDmodel}
\tau = r k_t i_m -  \tau_F \, .
\end{equation}
\looseness=-1

\subsection{Mechanical friction torque model}
\label{sec:cv_scv}
The friction torque $\tau_F$ characterizes the nonlinear friction behavior of the harmonic reducer, encompassing static friction, dynamic friction, break-away force, pre-sliding displacement, frictional lag or hysteresis, and stick-slip~\cite{marques2016survey}. Static friction occurs during the sticking phase when the sliding velocity is zero, while dynamic friction is predominant during the sliding phase. Break-away force refers to the force required to overcome stiction. The Stribeck effect describes the decrease or negatively sloped characteristic of friction force at low sliding velocities, transitioning from static to Coulomb friction.

The physics-based friction models are classified into two categories: static models such as Coulomb, Coulomb-viscous, and Stribeck, and dynamic models such as Dahl, LuGre, and generalized Maxwell-slip~\cite{olsson1998friction}. Detailed mathematical formulations of dynamic models are beyond the scope of this work.
\looseness=-1

The continuous Coulomb-viscous (CV) model is defined as
\begin{equation}
\label{eq:cv}
    \tau_F = k_c \tanh(k_a \dot{s}) + k_v \dot{s} \; ,
\end{equation}
where $k_c \tanh(k_a \dot{s})$ represents the smooth Coulomb friction~\cite{pennestri2016review}, $k_v \dot{s}$ represents viscous friction. $\tanh$ approximates the sign function without discontinuity, and $k_a$ determines determining how quickly the friction changes as the velocity changes.
The Stribeck-Coulomb-viscous (SCV) model integrates the Stribeck effect in the previous formulation~\cite{olsson1998friction} and is defined as
\begin{equation}
\label{eq:scv}
    \tau_F = k_v \dot{s} + k_c \tanh(k_a \dot{s}) + \left( k_s {-} k_c  \right) e^{-\abs{\frac{\dot{s}}{v_s}}^\alpha} \tanh(k_a \dot{s}) \; ,
\end{equation}
where $v_s$ is the Stribeck velocity, $k_s$ represents the breakaway friction, and $\alpha$ is an empirical parameter determining how fast the static friction component fades away as the velocity increases.
\looseness=-1

\subsection{Physics informed neural networks}
Physics-Informed Neural Networks (PINNs) are a specialized class of neural networks that incorporate physical laws directly into the learning process, providing a robust alternative to traditional data-driven models~\cite{abiodun2018state,cuomo2022scientific}. Unlike conventional NNs, which rely solely on observational data, PINNs leverage governing physics equations to guide training, allowing the model to infer accurate solutions even from sparse data. This is accomplished by embedding physical laws as constraints within the model, ensuring that the solutions adhere to established scientific principles. The training of a PINN involves minimizing a composite loss function that combines both data-driven and physics-based terms. Given a general differential equation $\mathcal{N}(\tau_F) = 0$, where $\mathcal{N}$ represents a differential operator and $\tau_F$ is the unknown solution, the total loss function $\mathcal{L}$ is formulated as
\begin{IEEEeqnarray}{cl}
    \IEEEnonumber
\mathcal{L} = \mathcal{L}_{\text{data}} + \mathcal{L}_{\text{physics}} &= \lambda_{\text{data}} \frac{1}{N} \sum_{i=1}^{N} \left( {\tau}_{F,\text{pred}}(\mathbf{x}_i) - {\tau}_{F,\text{true}}(\mathbf{x}_i) \right)^2 \\ &+ \lambda_{\text{physics}} \frac{1}{M} \sum_{j=1}^{M} \left( \mathcal{N}({\tau_F}_{\text{pred}}(\mathbf{x}_j)) \right)^2,
\label{eq:loss}
\end{IEEEeqnarray}
where $\mathcal{L}_{\text{data}}$ minimizes the difference between predicted and true values at data points $\mathbf{x}_i$, and $\mathcal{L}_{\text{physics}}$ ensures that the predicted solution ${\tau}_{F,\text{pred}}$ satisfies the governing physical laws at collocation points $\mathbf{x}_j$. Here, $N$ is the number of available data points, $M$ is the number of collocation points where the differential equation is evaluated, and $\lambda_{\text{data}},\lambda_{\text{physics}}$ are scaling parameters that balance the two loss components. PINNs have proven to be powerful tools across numerous domains, solving complex physical problems where traditional models often struggle due to insufficient data or the need to incorporate domain-specific knowledge.
\looseness=-1

\section{Methods}
\label{sec:methods}

This section details the friction identification procedure which starts with data acquisition from joint movements, followed by data processing to ensure the quality and relevance of the data. We adopt both static known models, such as CV and SCV, as well as a data-driven model, the PINN model. The aim is to accurately predict friction to improve joint torque control performance.
\looseness=-1

\begin{figure}[t]
    \centering
    \includegraphics[width=1.0\linewidth]{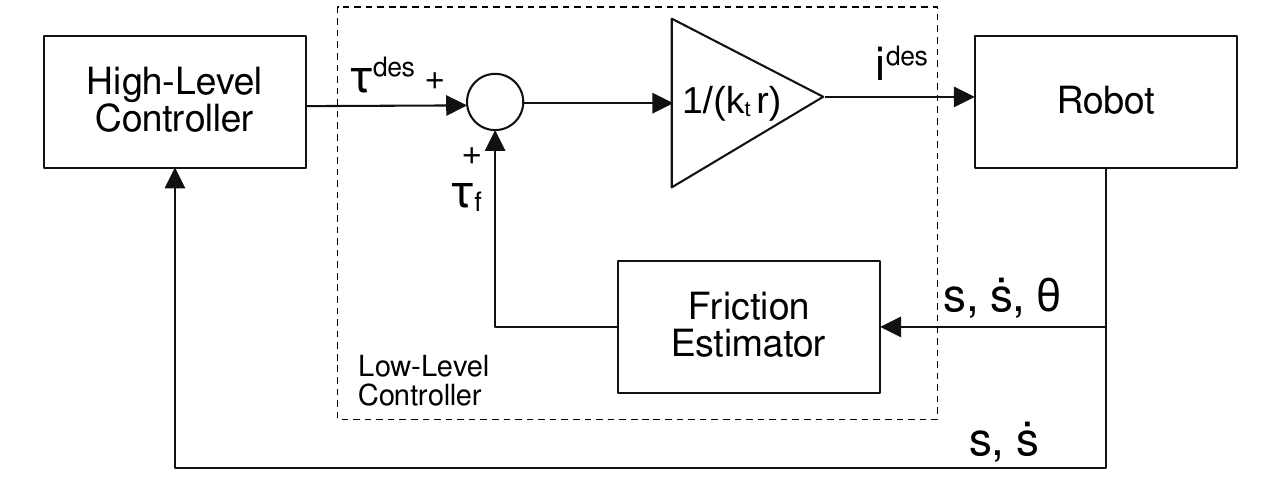}
    \caption{Block diagram of the two-layer controller architecture. The PINN implements the block Friction Estimator of the Low-Level controller.}
    \label{fig:controlarchitecture}
    \vspace{-20pt}
\end{figure}

\subsection{Data acquisition procedure}
The selection of the input excitation signal is a critical factor that determines the quality and accuracy of the estimated friction model, as well as the richness of the acquired data. A well-designed excitation signal captures the fundamental features of the system dynamics.
Experiments are conducted by generating desired motor current trajectories using a simple parameterized controller, with data acquired simultaneously on multiple joints. The sine waves exhibit increasing amplitudes and frequencies. Specifically, for each fixed amplitude value, we vary the frequency from its minimum to its maximum value. The minimum and maximum amplitudes and frequencies depend on the specific joint. Ramp signals are used with varying increments for each experiment. The trajectory is stopped when the motor current reaches the joint-dependent maximum value or when the joint position approaches its hardware limit. The final trajectory is represented by steps. Specifically in each experiment, we apply a step change in the motor current and each step is maintained for a few seconds. The trajectory is halted either after the specified duration or when the joint position approaches its hardware limit. To ensure that all potential static friction variations, which may arise from different initial configurations of the harmonic drive, are accounted for, each experiment is repeated multiple times starting from different initial joint configurations.
\looseness=-1

The generated motor currents are subsequently applied to a low-level PI motor current controller running at 20 KHz.
Data are logged at 500 Hz and then offline resampled at 1000 Hz for friction identification purposes. This resampling is necessary because the friction estimator is then integrated with a low-level joint torque controller running online at 1000 Hz on the robot computer.
\looseness=-1

\subsection{Data preprocessing}
Data pre-processing is essential to reduce measurement noise and outliers and generate joint velocities and accelerations. A low-pass Butterworth filter is employed to reduce the noise of measured motor currents and Inertia Measurement Units (IMU) signals~\cite{mahata2018optimal}. Joint velocities and accelerations are derived from collected joint positions using a Kalman Filter~\cite{welch1995introduction}. The tuning of these filters is straightforward, aiming to remove noise without losing important information.
Another critical data processing objective is to compute joint torques, which are not directly measured due to the absence of joint torque sensors on our humanoid robot. Joint torques can be computed based on the knowledge of the robot state and the robot model from Eq.~\ref{eq:robotdynamics}~\cite{traversaro2017thesis}. The friction torques, representing the ground truth ($\tau_{F, \text{true}}$) for the estimation algorithm, are computed from Eq.~\eqref{eq:HDmodel} as $\tau_{F, \text{true}} = r k_t i_m - \tau $ where the joint torque $ \tau $ is given by Eq.~\eqref{eq:robotdynamics}.
\looseness=-1

\subsection{PINN architecture}

\looseness=-1
The friction model is estimated using a Physics-Informed Neural Network (PINN). The joint velocity ($\dot{s}$) and the position error ($\Delta \theta = r s - \theta$) are buffered in a joint state history within a finite time window. The use of the position error ensures the model can differentiate between situations where the joint is stationary due to friction and cases where no movement is commanded, reflecting whether the motor intends to move or remain still. The PINN maps the joint state to a frictional torque value ($\tau_F$).
The loss function used to train the PINN combines both data-driven and physics-based components~\eqref{eq:loss} and is defined as
\begin{equation}
\begin{aligned}
&\mathcal{L}_{\text{data}} =  (1 - \lambda) \frac{1}{N} \sum_{i=1}^{N} \left( \tau_{F, \text{pred}} - \tau_{F, \text{true}} \right)^2 \\
&\mathcal{L}_{\text{physics}} = \lambda \frac{1}{M} \sum_{j=1}^{M} \left( \tau_{F, \text{pred}} - \tau_{F, \text{physics}} \right)^2 \; ,
\end{aligned}
\label{eq:combined_loss}
\end{equation}
where $\lambda \in [0, 1]$ is the regularization parameter, $\tau_{F, \text{physics}}$ is defined by~\eqref{eq:scv}, and $\tau_{F, \text{true}}$ is computed from \eqref{eq:HDmodel}.
\looseness=-1

This loss function ensures that the PINN predictions are accurate for the data and consistent with the underlying physics of the SCV model. The PINN consists of 2 hidden layers with \textit{ReLU} activation, incorporating dropout after each layer, and a linear output layer. We utilize the Weight and Biases (W\&B) platform to optimize the hyperparameters of our PINN model~\cite{wandb}. W\&B seamlessly integrates with TensorFlow and PyTorch, employing advanced techniques for hyperparameter optimization. We use the Random Search algorithm to select the set of hyperparameters from a specified search space~\cite{bergstra2012random}. Our objective is to identify the parameters yielding optimal performance by minimizing validation loss. Key parameters optimized include batch size, hidden layer sizes, learning rate, history length, regularization parameter $\lambda$, and dropout rate, with the number of epochs fixed.
\looseness=-1

\subsection{Controller architecture}
The control architecture is shown in Fig. \ref{fig:controlarchitecture} and is detailed below.
\looseness=-1

\subsubsection{High-Level Control}
\label{sec:HLC}
Based on the robot dynamics~\eqref{eq:robotdynamics}, the high-level control law computes the desired joint torques $\tau_{des}$ solving a constrained quadratic programming problem (QP). The implemented QP is a simplified version of the problem defined in~\cite{romualdi2022whole} and considers only the joint position regularization task in the control problem.
This task prevents huge variations of the desired joint accelerations and is described by \mbox{$\Psi_{s} = \ddot{s}^{*} - \begin{bmatrix} 0_{n \times 6}  & I_{n}  \end{bmatrix} \dot{\nu}$}. $\ddot{s}^{*}$ is defined as \scalebox{0.8}{$\ddot{s}^{*} = \ddot{s}^{des} + K_d (\dot{s}^{des} - \dot{s}) + K_p (s^{des} -s)$}, where $s^{des}$ is the desired joint trajectory, and $K_d$ and $K_p$ are two positive-defined diagonal matrices.
\looseness=-1

\begin{figure}[t]
    \centering
    \includegraphics[width=0.85\linewidth]{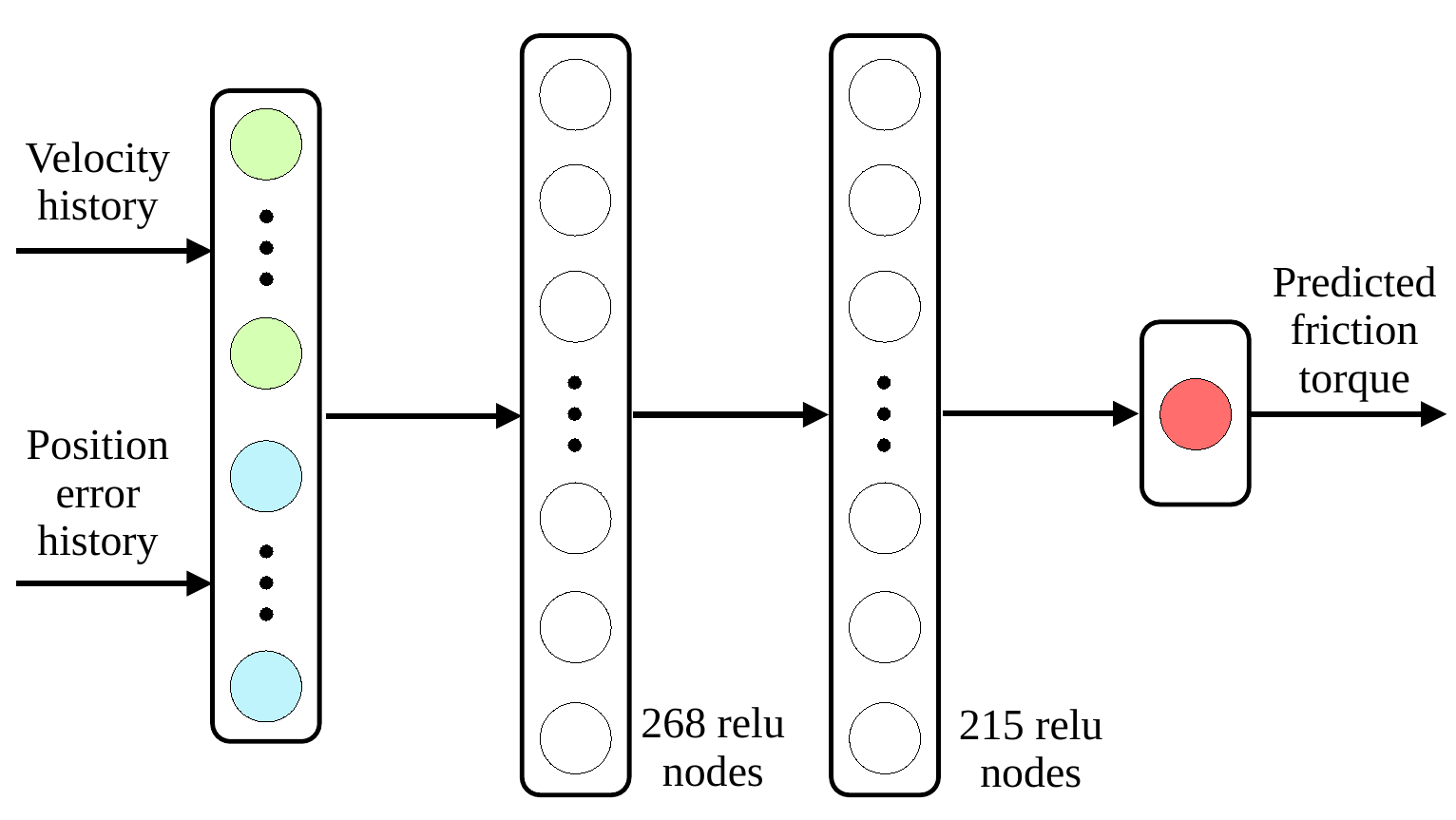}
    \caption{Neural Network architecture.}
    \label{fig:nnarchitecture}
    \vspace{-20pt}
\end{figure}

\subsubsection{Low-Level Torque Control}
The joint torques $\tau^{des}$ generated by the high-level controller described in Section \ref{sec:HLC}, are sent to the decentralized low-level joint torque controllers. The torque control schema is implemented as a feedforward controller including the compensation of the mechanical friction. The control law generates the reference currents to drive the motors.
\looseness=-1

\section{Results}
\label{sec:results}

\begin{figure}[t]
\centering
\begin{subfigure}{.98\columnwidth}
  \includegraphics[width=\textwidth]{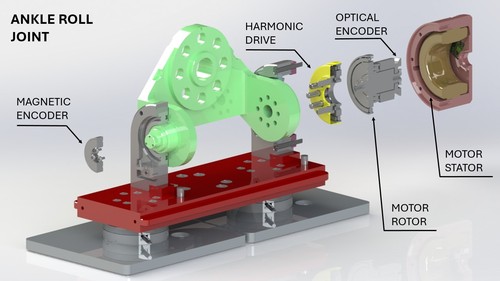}
\end{subfigure}%
\vspace{5pt} %
\begin{subfigure}{.98\columnwidth}
  \includegraphics[width=\textwidth]{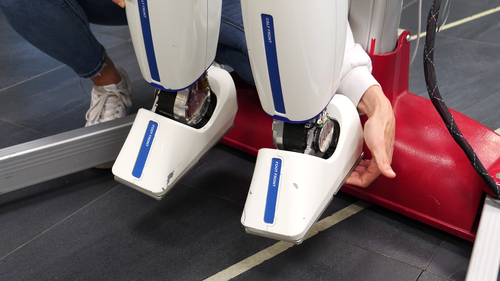}
\end{subfigure}
\caption{Exploded view of the ankle roll joint and left ankle roll while applying external disturbances.}
\label{fig:ankleext}
\vspace{-15pt}
\end{figure}

This section presents the validation of the joint friction torque estimation using the PINN and compares it with the CV and SCV models discussed in Section~\ref{sec:cv_scv}. We evaluate the performance of these models on the left ankle roll and left knee joints of the ergoCub humanoid robot (Fig. \ref{fig:ergocub} and ~\ref{fig:ankleext}). These joints are selected due to their distinct characteristics: the ankle roll joint has minimal load and is predominantly influenced by frictional forces, making friction compensation critical for precise control, while the knee joint experiences varying load conditions that affect friction differently depending on the leg configuration. We employ the high-level controller outlined in Section~\ref{sec:HLC} to track desired joint position trajectories. This controller generates the desired torque commands, which are then sent to the low-level joint torque controller. The low-level controller uses friction torque estimates from the friction models to provide compensation. The friction torque estimator is seamlessly integrated within the robot computer, operating automatically at a frequency of $1000$ Hz upon startup. Two distinct experiments are conducted to assess the effectiveness of the friction models, and the results are also showcased in the supplementary video. The code is available online~\cite{code}.
\looseness=-1

\subsection{CV and SCV models identification}
\label{sec:whitebox}

We utilized a data-driven approach to identify the parameters of the CV and SCV models using experimental data. The optimization process aims to minimize the mean squared error (MSE) between the predicted and measured friction torques,
\mbox{$\text{MSE} = \frac{1}{N} \sum_{i=1}^{N} ({\tau_{F, \text{pred}_i}} - {\tau_{F, \text{true}_i}})^2$}.
 The Adam optimizer, applied over $10000$ epochs, adjusts the model parameters such as $k_a$, $k_c$ and $k_v$ for the CV model~\eqref{eq:cv}, and $v_s$, $k_a$, $k_c$, $k_v$ and $\alpha$ for the SCV model~\eqref{eq:scv}. Both models used joint velocity data $\dot{s}$ as input and friction torque $\tau_F$ as the target output. Once trained, these parameters are employed for friction compensation in the low-level controller. The resulting parameters are listed in~\ref{table:cvscv}.
\looseness=-1

\subsection{PINN hyperparameter identification}

\begin{table}[t]
\centering
\caption{Identified parameters for the Coulomb-viscous (CV) and Stribeck-Coulomb-viscous (SCV) friction models.}
\setlength{\tabcolsep}{4pt}
\begin{tabular}{lcccc}
\hline
\textbf{Parameters} & $\text{CV}_{\text{ankle\_roll}}$ & $\text{SCV}_{\text{ankle\_roll}}$ & $\text{CV}_{\text{knee}}$ & $\text{SCV}_{\text{knee}}$ \\ 
\hline
$k_a$ [\SI{}{s\per rad}] & 10.53 & 2.78 & 50.8400 & 16.95 \\ 
$k_c$ [\SI{}{Nm}] & 1.2 & 1.0005 & 8.1 & 5.0 \\ 
$k_v$ [\SI{}{Nm.rad\per\second}] & 0.24 & 0.29 & 5.55 & 5.34 \\ 
$k_s$ [\SI{}{Nm}] & - & 6.0 & - & 9.7 \\ 
$v_s$ [\SI{}{rad\per\second}] & - & 0.13 &  -  & 5.4 \\ 
$\alpha$ & - & 0.6 & - & 0.5 \\
\hline
\label{table:cvscv}
\vspace{-15pt}
\end{tabular}
\end{table}
\looseness=-1

The hyperparameters of the PINN model are identified using the Weight and Biases tool~\cite{wandb}. Table~\ref{table:hyper} presents the optimal hyperparameters identified, with training taking approximately one hour and a half per run. The training loss for the ankle roll joint is $0.0044$, with a validation loss of $0.000878$, whereas for the knee joint, the training and validation losses are $0.323$ and $0.0787$, respectively.
\looseness=-1

\begin{figure*}[t]
\centering
\begin{subfigure}{.49\textwidth}
  \includegraphics[width=0.98\linewidth,center]{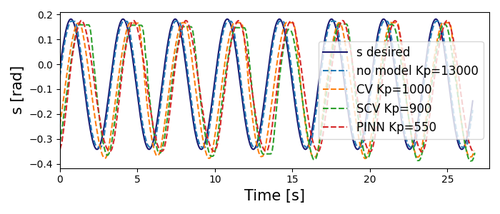}
  \vspace{-20pt}
  \caption{Joint position tracking.}
  \label{fig:kptrackingankle}
\end{subfigure}%
\hspace{\fill} %
\begin{subfigure}{.49\textwidth}
  \includegraphics[width=0.98\linewidth,center]{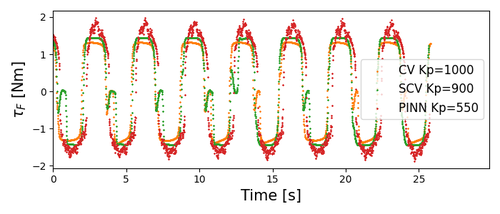}
  \vspace{-20pt}
  \caption{Estimated friction torques.}
  \label{fig:kpfrictionankle}
\end{subfigure}
\begin{subfigure}{.98\textwidth}
  \includegraphics[width=0.9\linewidth,center]{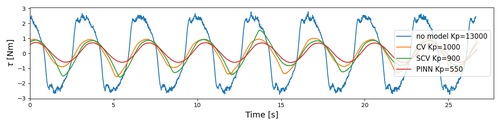}
  \vspace{-20pt}
  \caption{Desired joint torques.}
  \label{fig:kptorqueankle}
\end{subfigure}
\vspace{-5pt}
\caption{Left ankle roll results. Comparison of joint position tracking, friction estimation, and desired joint torques with different high-level control gains tuned on each specific friction model.}
\vspace{-10pt}
\end{figure*}
\begin{figure*}[t]
\centering
\begin{subfigure}{.48\textwidth}
  \includegraphics[width=0.98\linewidth,center]{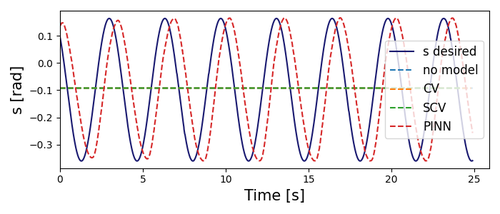}
  \vspace{-20pt}
  \caption{Joint position tracking.}
  \label{fig:kp550trackingankle}
\end{subfigure}%
\hspace{\fill} %
\begin{subfigure}{.48\textwidth}
  \includegraphics[width=0.98\linewidth,center]{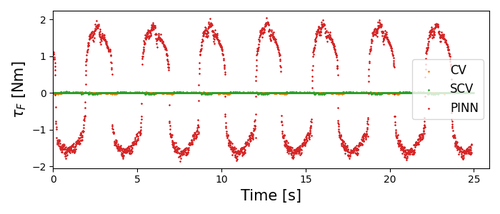}
  \vspace{-20pt}
  \caption{Estimated friction torques.}
  \label{fig:kp550frictionankle}
\end{subfigure}
\vspace{-5pt}
\caption{Left ankle roll results. (a)-(b) Comparison of the joint position tracking and estimated friction $K_p=550$ and $K_d=4$.}
\vspace{-15pt}
\end{figure*}

\begin{table}[t]
\centering
\caption{Physics Informed Neural Network hyperparameters identified using the Weight and Biase tool.}
\begin{tabular}{p{0.5\linewidth}cc}
\hline
\textbf{Hyperparameter} & \textbf{ankle\_roll} & \textbf{knee} \\
\hline
Epochs                  & 350  &  350    \\
Batch size              & 4316  &  4914   \\
Dropout rate            & 0.07  &  0.01176  \\
Hidden neurons layer 1     & 268  & 194  \\
Hidden neurons layer 2     & 215  & 247  \\
History length          & 20    &  22  \\
Learning rate           & 0.00076  & 0.00076 \\
$\lambda$               & 0.164  &  0.484 \\
Weight and biases initialization   & Random & Random \\
Optimizer & ADAM & ADAM \\
\hline
\label{table:hyper}
\end{tabular}
\vspace{-25pt}
\end{table}
\looseness=-1

\subsection{System performance with different friction models}
We evaluate the tracking performance of the friction models by having the left ankle roll and left knee joints follow a sinusoidal trajectory. The ergoCub's root link is fixed, allowing free motion of the legs. The high-level controller generates desired torque commands, which are transmitted to the feedforward low-level torque controller that integrates friction compensation from the respective models.

\begin{figure*}[t]
\centering
\begin{subfigure}{.49\textwidth}
  \includegraphics[width=0.98\linewidth,center]{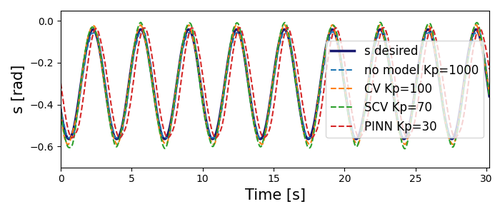}
  \vspace{-20pt}
  \caption{Joint position tracking.}
  \label{fig:trackingknee}
\end{subfigure}%
\hspace{\fill} %
\begin{subfigure}{.49\textwidth}
  \includegraphics[width=0.98\linewidth,center]{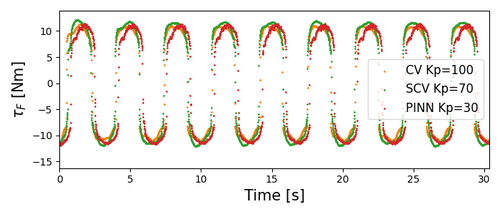}
  \vspace{-20pt}
  \caption{Estimated friction torques.}
  \label{fig:frictionknee}
\end{subfigure}
\begin{subfigure}{.98\textwidth}
  \includegraphics[width=0.9\linewidth,center]{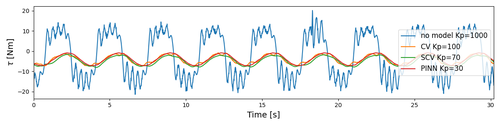}
  \vspace{-20pt}
  \caption{Desired joint torques.}
  \label{fig:torqueknee}
\end{subfigure}
\vspace{-5pt}
\caption{Left knee results. Comparison of joint position tracking, friction estimation, and desired joint torques with different high-level control gains tuned on each specific friction model.}
\vspace{-10pt}
\end{figure*}

We conduct repeated experiments using various friction models, including a case without friction compensation. To assess the impact of each friction model, we identified the minimum proportional gain $K_p$ required to track the desired trajectory accurately. The $K_d$ gain is set to $4$ and is never changed.  A lower $K_p$ indicates more effective friction compensation, allowing for precise tracking with minimal control effort. Fig.~\ref{fig:kptrackingankle} compares the performance of the high-level controller adopting the three different friction models on the ankle roll joint. The PINN model achieves the lowest $K_p$, followed by the SCV and CV models. Similar trends are observed for the knee joint, as depicted in Fig.~\ref{fig:trackingknee}. The superior performance of the PINN model in requiring the lowest $K_p$ underscores its effectiveness in accurately estimating and compensating for friction. Fig.~\ref{fig:kpfrictionankle} and \ref{fig:frictionknee} present the friction estimated during the experiments described above, while Fig.~\ref{fig:kptorqueankle} depicts the relationship between $K_p$ values and the corresponding desired torques. In Figure \ref{fig:kptorqueankle} we observe that higher $K_p$ values correspond to higher desired torques, reflecting increased control effort needed for precise joint movement. This highlights the PINN model's efficiency in minimizing control effort by accurate friction compensation. Conversely, the knee joint dynamics are primarily influenced by varying load rather than friction, making it less sensitive to $K_p$ adjustments (see Figure \ref{fig:torqueknee}). Overall, the significant disparity in desired torque values between no-compensation and friction-compensation cases is evident for both ankle roll and knee joints.
\looseness=-1

\begin{figure*}[t]
\centering
\begin{subfigure}{.48\textwidth}
  \includegraphics[width=0.98\linewidth,center]{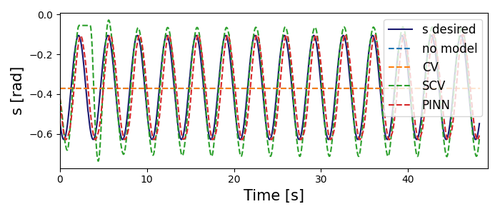}
  \vspace{-20pt}
  \caption{Joint position tracking.}
  \label{fig:kp30trackingknee}
\end{subfigure}%
\hspace{\fill} %
\begin{subfigure}{.48\textwidth}
  \includegraphics[width=0.98\linewidth,center]{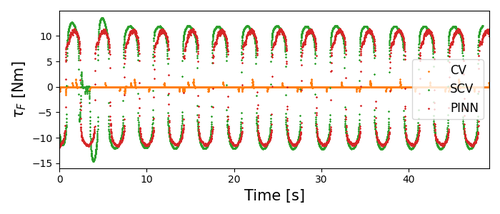}
  \vspace{-20pt}
  \caption{Estimated friction torques.}
  \label{fig:kp30frictionknee}
\end{subfigure}
\vspace{-5pt}
\caption{Left knee results. (a)-(b) Comparison of the joint position tracking and estimated friction with $K_p=30$ and $K_d=4$.}
\vspace{-15pt}
\end{figure*}

Fig.~\ref{fig:kptrackingankle} and Fig.~\ref{fig:trackingknee} show that systems with lower $K_p$, particularly those utilizing the PINN model, exhibit slight delays in joint position tracking, likely due to the less aggressive control action. In contrast, systems with higher gains, which lack proper friction compensation, generate higher desired torques and reduced tracking delays. Indeed, higher gains yield more aggressive responses, reducing the error between desired and actual joint positions. Introducing a feedback controller at the torque control level could combine the benefits of accurate friction compensation with improved tracking performance through feedback.
\looseness=-1

We also test the models using the lowest $K_p$ determined by the PINN-based friction estimation. The results show that the CV and SCV models are unable to generate sufficient torque to overcome static friction in the ankle roll joint, as shown in Fig.~\ref{fig:kp550trackingankle} and~\ref{fig:kp550frictionankle}. Conversely, for the knee joint, only the CV model fails to move the joint. However, the SCV model presents a higher tracking error for the sinusoidal trajectory compared to the PINN model, as evidenced in Fig.~\ref{fig:kp30trackingknee} and~\ref{fig:kp30frictionknee}, confirming the superior performance of the PINN model.
\looseness=-1

\subsection{Joint position stability and response to external wrenches}

\begin{table}[t]
\centering
\caption{Comparison of the results with different friction models for the left ankle roll joint.}
\setlength{\tabcolsep}{7pt}
\begin{tabular}{lcccc}
    \toprule
    Models    & RMSE [rad] & $\|\prescript{roll}{}{\mu_x}\|$ [Nm] & $K_p$ & $K_d$ \\
    \midrule
    No Model  & 0.009 & 21  & 13000 & 4 \\
    CV        & 0.05  & 4.7   & 1000 & 4 \\
    SCV       & 0.047 & 5.0   & 900 & 4  \\
    PINN      & 0.046 & 3.0   & 550 & 4  \\
    \bottomrule
\end{tabular}
\vspace{-12pt}
\label{tab:comparisonankle}
\end{table}
\begin{table}[t]
\centering
\caption{Comparison of the results with different friction models for the left knee joint.}
\setlength{\tabcolsep}{7pt}
\begin{tabular}{lcccc}
    \toprule
    Models    & RMSE [rad] & $\|\prescript{roll}{}{\mu_x}\|$ [Nm] & $K_p$ & $K_d$ \\
    \midrule
    No Model  & 0.0085 & 65.4  & 1000 & 4 \\
    CV        & 0.035  & 17.6   & 100 & 4 \\
    SCV       & 0.049 & 23.0   & 70 & 4  \\
    PINN      & 0.017 & 8.9   & 30 & 4  \\
    \bottomrule
\end{tabular}
\vspace{-20pt}
\label{tab:comparisonknee}
\end{table}
\looseness=-1
In our second experiment, we employ a high-level controller to maintain a fixed initial joint position under two distinct conditions: using the lowest $K_p$ values for all friction models and using $K_p$ values that demonstrated effective tracking in previous tests. This experiment aims to analyze the system response to an external disturbance and evaluate the efficacy of different friction models. As shown in Fig.~\ref{fig:ankleext}, the external disturbance is manually applied to attempt rotation of the joint about its rotation axis. This applied wrench is measured by two force/torque sensors positioned under the robot foot. The wrench is composed of three forces and three moments, but only one component directly influences the joint movement, while others are absorbed by the mechanical structure. To compute the disturbance wrench acting on the joint, we transform the measured wrench $\prescript{}{}{\mathrm{f}}_{measured}$ into a frame attached to the joint and whose \textit{x}-axis is aligned with the joint rotation axis, namely $f_{joint} = \prescript{joint}{}{X}_{measured} f_{measured}$. Finally, we extract from $f_{joint}$ only the moment about the \textit{x}-axis contributing to the joint rotation.
\looseness=-1

Using the $K_p$ values that provided good tracking in the previous tests, the results are summarized in Tables~\ref{tab:comparisonankle} and Table~\ref{tab:comparisonknee} for the two joints. The tables present the root mean square error (RMSE) of the joint position after the external disturbance ceased and the joint returned to the initial position. The tables also include the value of the external torque applied to the joint to move it from its initial position. 
\looseness=-1

When using the lowest $K_p=550$ for the ankle roll joint and $K_p=30$ for the knee joint, we observe that after the external disturbance is removed, the joint successfully returns to its starting position only when the PINN model is used as shown in Table~\ref{tab:comparisonlowkp}. For other friction models, the compensation is insufficient, and the desired torque generated by the controller is not enough to bring the joint back to its initial position. This indicates that the PINN model provides superior friction compensation, ensuring accurate joint position recovery even with a lower proportional gain.
\looseness=-1

\begin{table}[t]
\centering
\caption{Joint position tracking error after applying external disturbance while using different friction models and the same high-level controller gains.}
\setlength{\tabcolsep}{8pt}
\begin{tabular}{lcccc}
    \toprule
    & \textbf{No Model} & \textbf{CV} & \textbf{SCV} & \textbf{PINN} \\
    \midrule
    \textbf{Left Ankle Roll} & & & & \\
    RMSE [rad] & 0.19 & 0.2  & 0.21 & 0.046 \\
    $K_p=550$, $K_d=4$ & & & & \\
    \midrule
    \textbf{Left Knee} & & & & \\
    RMSE [rad] & 0.21  & 0.11   & 0.097 & 0.017 \\
    $K_p=30$, $K_d=4$ & & & & \\
    \bottomrule
\end{tabular}
\vspace{-15pt}
\label{tab:comparisonlowkp}
\end{table}

While a higher $K_p$ generally resulted in a lower RMSE, indicating better tracking accuracy, especially for the ankle roll joint, it also highlighted the drawbacks. Indeed, the lowest force is required with the lowest $K_p$. This observation underscores the benefits of accurate friction compensation. By accurately modeling and compensating for frictional forces, the joint requires less external torque to achieve desired movements. This not only enhances control precision but also minimizes mechanical stress and energy consumption, contributing to overall system efficiency and reliability.
\looseness=-1

\section{Conclusions}
\label{sec:conclusions}

The application of Physics-Informed Neural Networks (PINNs) for friction identification in humanoid robots offers a promising method, though there are considerations. Training directly on the robot may yield less accurate data than dedicated setups. However, utilizing existing sensor data and physical laws eliminates the need for additional hardware, which is advantageous. In addition, comprehensive validation across all joints is essential due to their differing friction characteristics, requiring specific hyperparameter tuning for each joint. PINNs present trade-offs in data needs, interpretability, and accuracy compared to traditional friction models, which can complicate their adoption in certain scenarios. Nonetheless, PINNs excel in capturing non-linear and dynamic friction effects that conventional models struggle with. Their scalability across multiple joints without rigid-body assumptions enhances their applicability in dynamic environments, facilitating more robust friction compensation and motion control in humanoid robots. Thus, while acknowledging these limitations, the benefits of employing PINNs underscore their potential to advance friction estimation capabilities in robotic systems.\looseness=-1

Future work should prioritize validating all joints of the ergoCub robot to ensure robustness across varying friction characteristics and refining joint-specific hyperparameter tuning. Additionally, we aim to integrate PINN models into joint torque control architectures for complex real-time applications, such as demonstrating their performance during the robot's walking scenarios.
\looseness=-1

\bibliography{IEEEexample}
\bibliographystyle{IEEEtran}

\end{document}